\pdfoutput=1

\documentclass[11pt]{article}

\newcommand{\version}{}   
\usepackage[\version]{acl}

\usepackage{times}
\usepackage{latexsym}

\usepackage[T1]{fontenc}

\usepackage[utf8]{inputenc}

\usepackage{multirow}
\usepackage{pdfpages}
\usepackage{hyperref}

\usepackage{color}
\usepackage{lipsum}
\usepackage{colortbl}

\def\parsum#1{\bgroup \textcolor{blue}{Paragraph summary: #1}\egroup}
\def\sectionsum#1{\bgroup \textcolor{green}{Section content: #1}\egroup \\}

\newcommand{\redacted}{\texttt{[redacted for anonymity]}}
\newcommand{\anonymity}[1]{\ifthenelse{\equal{\version}{review}}{\redacted}{#1}}

\usepackage{booktabs}

\newcommand{\thebook}{\emph{The Glannon Guide To Civil Procedure}}

\newcommand{\bert}{BERT}
\newcommand{\bertbs}{BERT-Base}
\newcommand{\lbert}{Legal-BERT}
\newcommand{\lbertbs}{Legal-BERT-Base}

\usepackage{framed}

\newif\ifarxiv
\arxivtrue

\title{The Legal Argument Reasoning Task in Civil Procedure}

\author{Leonard Bongard \and Lena Held \and Ivan Habernal\\
Trustworthy Human Language Technologies\\
Department of Computer Science\\
Technical University of Darmstadt\\
\texttt{leonard.bongard@stud.tu-darmstadt.de}\\
\texttt{\{lena.held, ivan.habernal\}@tu-darmstadt.de}
}

\begin{document}

\ifarxiv
\onecolumn
\noindent \textbf{The Legal Argument Reasoning Task in Civil Procedure}

\medskip
\noindent Leonard Bongard, Lena Held, Ivan Habernal

\bigskip
This is a \textbf{camera-ready version} of the article accepted for publication at the \emph{Natural Legal Language Processing Workshop 2022} co-located with EMNLP. The final official version will be published in the ACL Anthology later in 2022: \url{https://aclanthology.org/}

\medskip
Please cite this pre-print version as follows.
\medskip

\begin{verbatim}
@InProceedings{Bongard.et.al.2022.NLLP,
  title     = {{The Legal Argument Reasoning Task in Civil Procedure}},
  author    = {Bongard, Leonard and Held, Lena and Habernal, Ivan},
  booktitle = {Proceedings of the Natural Legal Language Processing
               Workshop 2022},
  pages     = {(to appear)},
  year      = {2022},
  address   = {Abu Dhabi, UAE}
}
\end{verbatim}
\twocolumn

\fi

\maketitle
\begin{abstract}
We present a new NLP task and dataset from the domain of the U.S. civil procedure. Each instance of the dataset consists of a general introduction to the case, a particular question, and a possible solution argument, accompanied by a detailed analysis of why the argument applies in that case. Since the dataset is based on a book aimed at law students, we believe that it represents a truly complex task for benchmarking modern legal language models. Our baseline evaluation shows that fine-tuning a legal transformer provides some advantage over random baseline models, but our analysis reveals that the actual ability to infer legal arguments remains a challenging open research question.
\end{abstract}

\section{Introduction}

Arguing a legal case is an essential skill that aspiring lawyers must master. This skill requires not only knowledge of the relevant area of law, but also advanced reasoning abilities, such as using analogy arguments or finding implicit contradictions. Despite recent significant contributions aimed at setting objective benchmarks for modern NLP models in various areas of legal language understanding \cite{Chalkidis.et.al.2022.ACL}, there is still no complex task dealing with argument reasoning in legal matters.

In this paper, we propose a new task and provide a new benchmark dataset. We believe that a genuinely difficult task, coming from legal education, will help to demonstrate both the capabilities and the limitations of the current state-of-the-art legal transformation models, such as \lbert\ \citep{chalkidis2020legal}. In particular, we present a new, publicly available\footnote{See the Data sheet in the appendix for details.} legal corpus for binary text classification of U.S. civil procedure problems. The goal is to classify whether a solution to a given question is correct or incorrect.
The data for the corpus is based on the \thebook\ by Joseph Glannon \cite{Glannon.2018}, which is aimed at law students. The book allows for the study of basic U.S. civil procedure topics and also contains multiple-choice questions on civil procedure problems to test the reader.

With this newly created corpus, we also intend to investigate the performance of the different approaches and establish baselines and an error analysis. All source codes used to parse, extract and reformat the data and evaluate the solution methods can be found at 
\anonymity{\url{https://github.com/trusthlt/legal-argument-reasoning-task}}.

\section{Related work}
\paragraph{General QA and argument reasoning benchmarks}
Although the landscape of Question Answering (QA) corpora is vast, there are several categories and nuances which enable a more fine-grained division. For a better overview of the field, we refer the reader to a recent survey~\citep{Rogers.et.al.2022.ACM}. A possible distinction can be made on the basis of the target skill to be learned, among which commonsense reasoning is a more challenging one. In order to contribute to the learning of reasoning, a corpus must be designed in such a way that questions cannot be answered with a given context or linguistic cues alone. Several corpora take this into account by making their QA design ``hard to answer'' without additional context~\citep{mostafazadeh-etal-2017-lsdsem, huang-etal-2019-cosmos}.
To further support the ability to reason, some datasets provide an explanation in addition to the typical format of context, question and answer~\citep{jansen-etal-2016-whats, Camburu.et.al.2018.NeurIPS, lamm-etal-2021-qed}. Somewhat specific is the Argument Reasoning Comprehension Task by \citet{Habernal.et.al.2018.NAACL} in which the goal is to choose one of two contradicting warrants that connect a premise with the given claim.
Apart from traditional commonsense reasoning which requires general knowledge and understanding~\citep{talmor-etal-2019-commonsenseqa, sap-etal-2019-social}, there are also datasets in specialized fields like biomedical QA~\citep{Tsatsaronis.et.al.2015.BMC} which target domain specific factoid knowledge.
\paragraph{Legal question answering and legal reasoning}
In legal NLP, the number of publicly available corpora is considerably smaller, especially in the areas of argumentation and reasoning.
An early compilation of legal questions in a multiple-choice format is found in~\citep{Fawei.et.al.2016.LREC}. Each of the 100 questions taken from the United States Multistate Bar Examination is interpreted as an entailment task with one correct answer (entailment) and three incorrect answers (non-entailment). However, their evaluation shows that a mere similarity between theory (question) and hypothesis (answer) is insufficient to solve this task. A follow-up work extends the bar exam corpus with older exam practice questions and reformulates the task into an ``Answer-Sentence-Selection'' task instead of textual entailment~\citep{John.et.al.2017.ASAIL}. Additionally, 15 of the questions are annotated with an explanation.
Pursuing a different approach, \citet{Holzenberger.et.al.2020.NLLP} try to answer natural language questions in the domain of tax law by encoding knowledge as a set of rules with the help of a prolog solver. \citet{Holzenberger.et.al.2021.IJCNLP} extend the previously introduced corpus and subdivide statutory reasoning into a sequence of smaller tasks. Although this logic-based approach facilitates the answering, instantiation from natural language is not a trivial task.
Apart from resources in English, \citet{Zhong.et.al.2020.AAAI} compile a corpus for Legal QA in Chinese with knowledge-driven questions and case-analysis questions from the National Judicial Examination of China. They identify different types of reasoning needed to answer the questions and conclude that existing models lack reasoning ability, especially for knowledge-driven questions.
Contests such as the COLIEE competition~\citep{Kano.et.al.2019.COLIEE, Rabelo.et.al.2022.COLIEE} also include legal question answering tasks in a format that requires retrieval of relevant articles and entailment. 
The corpus for these tasks is taken from Japanese Bar Exams and manually translated into English.    
Although there are several resources available that deal with legal question answering, only few of them target statutory law in the English language. There is also still a shortcoming of datasets dedicated to argumentative answers in Legal QA which we aim to complement with our work.

\section{Dataset construction}

We built the dataset based on the content of \thebook\ \citep{Glannon.2018}. We collaborated with the author and the publisher\footnote{Joe Terry, vice president of Aspen Publishing} and negotiated a permission agreement under which the resulting dataset will be freely available to the research community.

The book contains 25 chapters with multiple choice questions.
Each chapter deals with a specific topic and asks and answers several questions.
The topic of a question is introduced beforehand in an introduction.
Each question is followed by three to five answer candidates, where only one candidate is correct.
The answer candidates can be considered as hypotheses.
Choosing the correct answer requires examining if the various prerequisites of a hypothesis hold.
Whether an answer is correct or incorrect is then discussed in the analysis afterward.

We parsed the book first fully automatically, as the structure allowed us to extract the components of each instance in the resulting dataset (introduction, question, answers, analysis).
Anomalies in the structure, e.g., the same introduction for two questions, were caught by additional parsing rules.
However, minor portions of the book had to be extracted manually, for instance
the correctness of the answer candidate because
the solution is addressed within the analysis' free text.
The analysis is loosely designed as follows. Each paragraph deals with an answer candidate and classifies it as true or false. Therefore, we decided to further separate the analysis to isolate the relevant aspect for each answer.
There were no keywords or structure artifacts indicating where to split the text. 
Furthermore, several inconsistencies regarding the structure exist.
Thus, separating the analysis had to be done manually too. Two complete examples (one incorrect and one correct, labeled as $0$ and $1$, respectively) are shown in Appendix~\ref{app:ex1}.

Separating the analysis allows the creation of a binary classification task, which should be suitable for many application scenarios.
The final legal argument reasoning task is defined as follows:
\begin{description}
    \item[Task] Given a question with a possible correct answer and a short introduction to the topic of the question, identify if the answer candidate is correct or incorrect.
\end{description}
After parsing the book, each question and answer pair is ordered as follows: \textit{1. Question; 2. Answer; 3. Solution; 4. Analysis; 5. Complete Analysis; 6. Introduction}.

\citet{Glannon.2018} intended to ask more difficult questions at the end of each chapter.
We thus created a data split accordingly.
The \textit{rational data split} divides the first 80\% of questions of each chapter into the train set, the following 10\% of questions into the dev set, and the last 10\% (which tend to be more difficult) into the test set.

The final dataset consists of 918 entries.
To evaluate if there is any evidence that some answers were more likely to be the correct answer than others, a distance analysis using \textit{SentenceBert}\footnote{\url{https://huggingface.co/sentence-transformers/all-mpnet-base-v2}}~\citep{reimers-gurevych-2019-sentence} was applied.
This evaluation method was applied to the train-, dev- and test-set.
The results indicate that there is little to no evidence that clues exist in the train and dev set.
The test set shows an increased accuracy at guessing the correct answer, which is 9.88\% more than the expected result of 24.5\% (calculated average of guessing the correct answer of a question).

\section{Baseline experiments}

A first baseline evaluation of the task included classification through pre-trained transformer models.
To evaluate the performance, the macro $F_1$ score is used. 
We chose \lbert\ as the classification model because of its legal tech application domain (although the model was not pre-trained with American civil procedure data).
We evaluated \bert\ \cite{devlin2018bert} as well as \lbert\ \cite{chalkidis2020legal} with and without fine-tuning giving it the question, answer, and introduction on an instance as input.
Another evaluation with these models included only the question and answer as input.
For fine-tuning we experimented as suggested by \citet{chalkidis2020legal} with the learning-rate, weight-decay and the dropout-rate.
We finished training through early-stopping (patience was set to 3).
The deep learning approach uses two techniques to bypass the maximum token limit problem.
\begin{description}
    \item[Sliding Window Simple (SWS)] separates the concatenated question and introduction into chunks. Each chunk is then classified, and the result is the average of the predicted outputs.
    \item[Sliding Window Complex (SWC)] divides the introduction into multiple chunks, where each chunk contains the complete question and is padded up with the introduction. Each chunk is then classified where the result is the average of the predicted outputs.
\end{description}
Table~\ref{tab:dl.results} displays the best results out of 15 runs with different hyper-parameters.
The fine-tuned \lbert\ model performs best with the SWC approach and outperforms the best performing \bert\ model as well as the random baseline significantly.
Furthermore, there is a notable difference between the performance of \bert\ and \lbert\ using the SWS method.

\begin{table}[]
    \centering
    \begin{tabular}{ lrr}
        \toprule
        Classifier & Accuracy & $F_1$ \\
        \midrule
        Random Baseline & 50.33 & 46.74 \\
        \vspace{0.5em}Majority Baseline & 80.52 & 44.21 \\
        \lbertbs\ & & \\
        --- (q,a) & 68.86 & 50.39 \\
        --- SWS (q,e,a) & 61.54 & 45.19 \\
        \vspace{0.5em}--- SWC (q,e,a) & 62.63 & 49.83 \\
		--- Finetuned (q,a) & 80.22 & 44.51 \\ 
        --- Finetuned SWS (q,e,a) & 81.31  & 63.03 \\
        \vspace{0.5em}--- Finetuned SWC (q,e,a) & 76.92  & \textbf{65.73} \\
        \bertbs\ & & \\
        --- Finetuned (q,a) & 80.22 & 44.51 \\
        --- Finetuned SWS (q,e,a) & 71.43  & 50.71 \\
        --- Finetuned SWC (q,e,a) & 80.22  & 56.80 \\
    \bottomrule
    \end{tabular}
    \caption{Accuracy and macro $F_1$-score (in \%) of transformer based models on the test set. To fit the complete question and introduction, the Sliding Window Simple (SWS) and Sliding Window Complex (SWC) are used. \label{tab:dl.results}}
    
\end{table}

\section{Analysis and discussion}
\begin{figure}[t!]
    \centering
    \begin{framed}
    \begin{description}
    \item[\textbf{Question:}] 14. Additions and objections. In July 2006, a week before the three-year statute of limitations passes, Carson sues Herrera in federal court for breach of a contract to design a computer system for his store in Calpurnia, Illinois. In July 2007, he moves to amend his complaint to add a claim for violation of the state Consumer Protection Act, based on the same dispute. The Consumer Protection Act has a two-year statute of limitations.
    \item[\textbf{Answer 1:}]The second claim would not be barred by the limitations period, as long as the judge grants the motion to amend.
    \item[\textbf{Answer 2:}]The second claim would ‘‘relate back’’ to the date of the original filing of the case, and therefore would not be barred by the statute of limitations.
    \item[\textbf{Answer 3:}]The second claim will be barred by the limitations period, because it will not ‘‘relate back’’ to the original filing under Rule 15.
    \item[\textbf{Answer 4:}]\textit{The amendment will be barred, even if it relates back to the filing of the original complaint.}
    \end{description}
	\end{framed}
    \caption{\label{fig:sample.predicted} The fine-tuned \lbert\ model predicts every possible answer of the corresponding question as correct. However, only answer 4 (italic) is correct.}
\end{figure}
Understanding legal argumentation is not an easy task by any means. Therefore it is not surprising that the performance of the transformer model is struggling. 
\thebook\ is an educational book to help students learn civil procedure questions.
Thus, even professionals have problems answering case law questions. 
A comparison to a human baseline would be beneficial to evaluate the overall performance of our model. 
We leave establishing the human upper bound for future work.

We did a brief error analysis by comparing the classification results between the fine-tuned \bert\ and \lbert\ model.
Out of 91 samples, the \bert\ model labels 6 of them as correct, while the \lbert\ model labels 21 as correct.
The \bert\ model predicted 3 of the 18 correct samples correctly, the \lbert\ model predicted 9 of them correctly.
17 answers have divided model predictions.
We read these samples for the error analysis, to understand the prediction.
We assume that the legal language used in the data the \lbert\ model is pretrained on, has an impact on the prediction results. 
This could be additionally indicated by the low amount of samples which are labeled as correct by the \bert\ model.
We further noticed that some questions have multiple answers that the \lbert\ model considers correct, even though the assertion of these answers differs. 
One example can be seen in Figure \ref{fig:sample.predicted}.

In an attempt to understand why the fine-tuned model has labeled all answers as correct we tried to follow the classification process through the usage of Captum, a Pytorch model interpretability library \cite{kokhlikyan2020captum}. 
Captum is used to calculate the attribution of each word vector as input feature for the final prediction.
However, as visualized in appendix \ref{Appendix:ErrorAnalyis}, a similar pattern for labeling each answer could not be found, despite the similarity of vocabulary between the answer possibilities.
Inconsistencies like these reveal that the \lbert\ model does not comprehensively reason about the answer.

Another explanation for the shortcomings of the evaluated models could be their inherent structure. In our dataset, concatenating the question, answer and introduction leads to 689 (646, 835) words or 3508 (3243, 4245) characters on average for the rational data split. Although the Sliding Window methods mitigate the token limits of BERT, a model that can deal with longer documents, like Longformer~\citep{Beltagy2020Longformer} or Big Bird~\citep{Zaheer.2020.NEURIPS} could prove to be more efficient. While a pretrained version of the Longformer architecture based on legal input exists in Chinese~\citep{Xiao.2021.Lawformer}, to the best of our knowledge there is no English equivalent available. The computationally expensive pretraining of such a model and testing it with our new dataset is left for future work.

Even more so, we have not included the most distinguishing property of our dataset in the experiments: the analysis. It is used to explain in human language why the answer to a question is correct or incorrect. As a possible future task, it would be interesting to see if this explanation could be used to boost the reasoning capabilities of a model.

\section{Conclusion}

We present a new challenging NLP task whose solution requires deeper knowledge and reasoning skills. We compare multiple transformer baselines and provide an error analysis showing that the correct prediction of the model for one instance does not prevent incorrect predictions for other relevant instances. We have obtained a license to share the dataset from the author of the original book and its publisher, and hope that it will help advance research in the complex field of legal argument reasoning.

\section*{Acknowledgements}
We would like to thank John Glannon and Aspen Publishing for their support.
The independent research group TrustHLT is supported by the Hessian Ministry of Higher Education, Research, Science and the Arts. This work has been partly funded by the German Research Foundation as part of the ECALP project (HA 8018/2-1).

\bibliography{bibliography}

\appendix

\section{Incorrect Example}
\label{app:ex1}

\paragraph{Introduction}
My students always get confused about the relationship between removal to federal court and personal jurisdiction. Suppose that a defendant is sued in Arizona and believes that she is not subject to personal jurisdiction there. Naturally, she should object to personal jurisdiction. But suppose further that she isn’t sure that her objection will carry the day; it’s a close issue, as so many personal jurisdiction issues are under minimum contacts analysis. And suppose that her tactical judgment is that, if she must litigate in Arizona, she would rather litigate in federal court in Arizona. What should she do? One thing she could do is to move to dismiss in the Arizona state court. But that motion is not likely to be ruled on within thirty days, and if she’s going to remove, she’s got to do it within thirty days. So she could do one of two things: She could move to dismiss for lack of personal jurisdiction in the state court (assuming that the state rules allow her to do that) and then remove to federal court within the thirty-day period. Her motion would then be pending in federal court instead of state court, and the federal court would rule on it. Or she could remove the case before a response was due in state court, and then, after removal, raise her objection to personal jurisdiction by a Rule 12(b)(2) motion to dismiss or in her answer to the complaint in federal court. Either way, the point is that removal does not waive the defendant’s right to object to personal jurisdiction. It simply changes the court in which the objection will be litigated. It is true that, after removal, the question will be whether the federal court has personal jurisdiction. But generally the scope of personal jurisdiction in the federal court will be the same as that of the state court, because the Federal Rules require the federal court in most cases to conform to state limits on personal jurisdiction. Fed. R. Civ. P. 4(k)(1)(A). I’ve stumped a multitude of students on this point. Consider the following two cases to clarify the point.
\paragraph{Question}
7. A switch in time. Yasuda, from Oregon, sues Boyle, from Idaho, on a state law unfair competition claim, seeking \$250,000 in damages. He sues in state court in Oregon. Ten days later (before an answer is due in state court), Boyle files a notice of removal in federal court. Five days after removing, Boyle answers the complaint, including in her answer an objection to personal jurisdiction. Boyle’s objection to personal jurisdiction is
\paragraph{Answer}
not waived by removal, but will be denied because the federal courts have power to exercise broader personal jurisdiction than the state courts.
\paragraph{Solution}
0
\paragraph{Analysis}
C is also wrong, because it suggests that, after removal, personal jurisdiction over Boyle will be tested by a different standard from that used in state court. In a diversity case, the reach of the federal court’s personal jurisdiction is governed by Fed. R. Civ. P. 4(k)(1)(A), which provides that the defendant is subject to personal jurisdiction in the federal court if she ‘‘is subject to the jurisdiction of a court of general jurisdiction in the state where the district court is located.’’ In other words, if the state courts of Oregon could exercise jurisdiction over Boyle, the Oregon federal court can; otherwise not.
\paragraph{Complete Analysis}
There are so many ways to go astray on this issue that I had to include five choices . . . and I could have made it seven! Surely the farthest astray is E. The fact that the court has subject matter jurisdiction over this diversity case does not mean that it has personal jurisdiction over Boyle. Though easily confused, the subject matter and personal jurisdiction analyses are separate; the court must have both subject matter jurisdiction over the case and personal jurisdiction over the defendant in order to proceed with the case. A reflects the faulty assumption that removal waives objections to personal jurisdiction. It doesn’t; it simply changes the forum in which the personal jurisdiction question will be litigated. Boyle may remove the case, and then respond to it, raising his defenses and jurisdictional objections in federal court. And B is wrong, because Boyle removed the case before the answer was due in state court. It is true, under some states’ procedural rules, that answering a complaint without including an objection to personal jurisdiction would waive it. But where a defendant removes before a response is due in state court, she does not waive any defenses by removal. She simply changes the forum in which such defenses will be raised. See Fed. R. Civ. P. 81(c)(1) (Federal Rules govern procedure after removal). C is also wrong, because it suggests that, after removal, personal jurisdiction over Boyle will be tested by a different standard from that used in state court. In a diversity case, the reach of the federal court’s personal jurisdiction is governed by Fed. R. Civ. P. 4(k)(1)(A), which provides that the defendant is subject to personal jurisdiction in the federal court if she ‘‘is subject to the jurisdiction of a court of general jurisdiction in the state where the district court is located.’’ In other words, if the state courts of Oregon could exercise jurisdiction over Boyle, the Oregon federal court can; otherwise not. D is the correct answer. Boyle has not waived his objection to personal jurisdiction. If the federal court lacks jurisdiction over Boyle, it should dismiss the case, even though it was properly removed. Now, another.

\section{Correct Example}
\label{app:ex2}

\paragraph{Question}
7. A switch in time. Yasuda, from Oregon, sues Boyle, from Idaho, on a state law unfair competition claim, seeking \$250,000 in damages. He sues in state court in Oregon. Ten days later (before an answer is due in state court), Boyle files a notice of removal in federal court. Five days after removing, Boyle answers the complaint, including in her answer an objection to personal jurisdiction. Boyle’s objection to personal jurisdiction is
\paragraph{Answer}
not waived by removal. The court should dismiss if there is no personal jurisdiction over Boyle in Oregon, even though the case was properly removed.
\paragraph{Solution}
1
\paragraph{Analysis}
D is the correct answer. Boyle has not waived his objection to personal jurisdiction. If the federal court lacks jurisdiction over Boyle, it should dismiss the case, even though it was properly removed.
\paragraph{Complete Analysis}
There are so many ways to go astray on this issue [...]. \emph{Same as in Appendix~\ref{app:ex1}}.
\paragraph{Introduction}
My students always get confused about the relationship between removal to federal court and personal jurisdiction. [...] \emph{Same as in Appendix~\ref{app:ex1}}.

\section{Error Analyis}\label{Appendix:ErrorAnalyis}
\begin{figure*}[h]
    \centering
    \includegraphics[width=\linewidth]{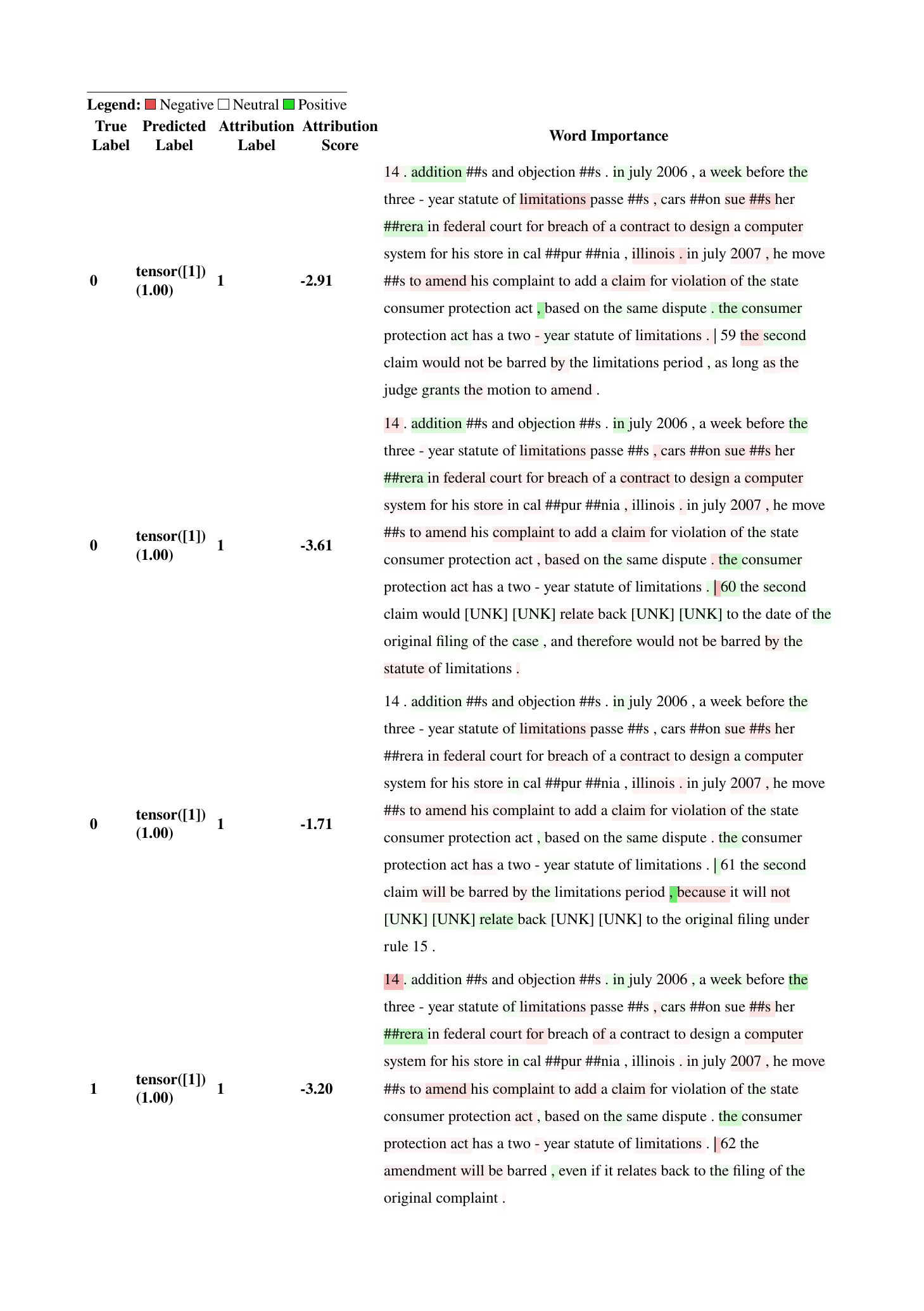}
    \caption{\lbert\ Model interpretability with Captum:}
\end{figure*}

\clearpage

\section{Data sheet} \label{Appendix:Datasheet}
The data sheet is provided following a template\footnote{\url{https://www.overleaf.com/latex/templates/datasheet-for-dataset-template/jgqyyzyprxth}} for \textit{Datasheets for datasets}~\citep{datasheet}. We have answered the questions to the best of our knowledge, but we would like to note that we can only make reliable statements about our collection process of the data but not the original book.

\definecolor{darkblue}{RGB}{46,25, 110}

\newcommand{\dssectionheader}[1]{%
   \noindent\framebox[\columnwidth]{%
      {\fontfamily{phv}\selectfont \textbf{\textcolor{darkblue}{#1}}}
   }
}

\newcommand{\dsquestion}[1]{%
    {\noindent \fontfamily{phv}\selectfont \textcolor{darkblue}{\textbf{#1}}}
}

\newcommand{\dsquestionex}[2]{%
    {\noindent \fontfamily{phv}\selectfont \textcolor{darkblue}{\textbf{#1} #2}}
}

\newcommand{\dsanswer}[1]{%
   {\noindent #1 \medskip}
}

\dssectionheader{Motivation}

\dsquestionex{For what purpose was the dataset created?}{Was there a specific task in mind? Was there a specific gap that needed to be filled? Please provide a description.}

\dsanswer{The dataset was created to enable research on reasoning towards civil procedure legal arguments. The dataset was created intentionally with that task in mind, focusing on the content provided by the book \thebook\ containing civil procedure problems.
}

\dsquestion{Who created this dataset (e.g., which team, research group) and on behalf of which entity (e.g., company, institution, organization)?}

\dsanswer{The dataset was created by \anonymity{Leonard Bongard, Lena Held, and Ivan Habernal (Technical University Darmstadt, Germany)}, based on a book by Joseph Glannon (Suffolk University, USA).
The creators of the dataset had no influence on the creation and publication of the book. 
The correctness of the solutions lies solely with the author and publisher of the book.
}

\dsquestionex{Who funded the creation of the dataset?}{If there is an associated grant, please provide the name of the grantor and the grant name and number.}

\dsanswer{N/A}

\dsquestion{Any other comments?}

\dsanswer{None.
}

\bigskip
\dssectionheader{Composition}

\dsquestionex{What do the instances that comprise the dataset represent (e.g., documents, photos, people, countries)?}{ Are there multiple types of instances (e.g., movies, users, and ratings; people and interactions between them; nodes and edges)? Please provide a description.}

\dsanswer{The instances are civil procedure problems extracted from the book \thebook. Multiple topics of civil procedure are covered in the book and are represented through the instances.
}

\dsquestion{How many instances are there in total (of each type, if appropriate)?}

\dsanswer{There are 918 instances in total. Each question-answer pair is treated as a separate instance.
}

\dsquestionex{Does the dataset contain all possible instances or is it a sample (not necessarily random) of instances from a larger set?}{ If the dataset is a sample, then what is the larger set? Is the sample representative of the larger set (e.g., geographic coverage)? If so, please describe how this representativeness was validated/verified. If it is not representative of the larger set, please describe why not (e.g., to cover a more diverse range of instances, because instances were withheld or unavailable).}

\dsanswer{The dataset contains all possible instances.
}

\dsquestionex{What data does each instance consist of? “Raw” data (e.g., unprocessed text or images) or features?}{In either case, please provide a description.}

\dsanswer{Each instance consists of a question, a corresponding answer, a solution, an analysis of the specific answer, the complete analysis of the question, and a topic introduction. The data is not further processed.
}

\dsquestionex{Is there a label or target associated with each instance?}{If so, please provide a description.}

\dsanswer{The label is the correctness or incorrectness of the answer derived from the analysis in binary format (0 or 1).
}

\dsquestionex{Is any information missing from individual instances?}{If so, please provide a description, explaining why this information is missing (e.g., because it was unavailable). This does not include intentionally removed information, but might include, e.g., redacted text.}

\dsanswer{The book contains multiple choice questions that were parsed into a binary classification format. However, there exist answers like "None of the answers are correct" which are excluded in our dataset. These answers cannot be used with our approach for reasoning.
}

\dsquestionex{Are relationships between individual instances made explicit (e.g., users’ movie ratings, social network links)?}{If so, please describe how these relationships are made explicit.}

\dsanswer{No.
}

\dsquestionex{Are there recommended data splits (e.g., training, development/validation, testing)?}{If so, please provide a description of these splits, explaining the rationale behind them.}

\dsanswer{The author of the book intended to ask more difficult questions at the end of each chapter.
Thus, we created a data split accordingly.
The \textit{rational data split} divides the first \textit{80\%} of questions of each chapter into the train set, the following \textit{10\%} of questions into the dev set, and the last \textit{10\%} into the test set. 
}

\dsquestionex{Are there any errors, sources of noise, or redundancies in the dataset?}{If so, please provide a description.}

\dsanswer{Since each question has multiple possible answers and each answer is assigned to a separate instance, there are redundancies in the content of the question, the complete analysis, and the explanation. For each instance, the analysis is also contained in the complete analysis.
}

\dsquestionex{Is the dataset self-contained, or does it link to or otherwise rely on external resources (e.g., websites, tweets, other datasets)?}{If it links to or relies on external resources, a) are there guarantees that they will exist, and remain constant, over time; b) are there official archival versions of the complete dataset (i.e., including the external resources as they existed at the time the dataset was created); c) are there any restrictions (e.g., licenses, fees) associated with any of the external resources that might apply to a future user? Please provide descriptions of all external resources and any restrictions associated with them, as well as links or other access points, as appropriate.}

\dsanswer{The dataset is self-contained. However, answering the questions requires an understanding of US civil procedure, which may change over time.
}

\dsquestionex{Does the dataset contain data that might be considered confidential (e.g., data that is protected by legal privilege or by doctor-patient confidentiality, data that includes the content of individuals non-public communications)?}{If so, please provide a description.}

\dsanswer{No.
}

\dsquestionex{Does the dataset contain data that, if viewed directly, might be offensive, insulting, threatening, or might otherwise cause anxiety?}{If so, please describe why.}

\dsanswer{Some instances discuss civil procedure cases, which may discuss socially relevant issues like discrimination or racism.
}

\dsquestionex{Does the dataset relate to people?}{If not, you may skip the remaining questions in this section.}

\dsanswer{Unknown to the authors of the datasheet. 
}

\dsquestionex{Does the dataset identify any subpopulations (e.g., by age, gender)?}{If so, please describe how these subpopulations are identified and provide a description of their respective distributions within the dataset.}

\dsanswer{No. 
}

\dsquestionex{Is it possible to identify individuals (i.e., one or more natural persons), either directly or indirectly (i.e., in combination with other data) from the dataset?}{If so, please describe how.}

\dsanswer{It is possible to identify some individuals indirectly by the occurrence of a name in a precedent. By looking up the precedent in an external source, a natural person can be inferred. (e.g. Swift v. Tyson)
}

\dsquestionex{Does the dataset contain data that might be considered sensitive in any way (e.g., data that reveals racial or ethnic origins, sexual orientations, religious beliefs, political opinions or union memberships, or locations; financial or health data; biometric or genetic data; forms of government identification, such as social security numbers; criminal history)?}{If so, please provide a description.}

\dsanswer{No.
}

\dsquestion{Any other comments?}
\dsanswer{None.
}

\bigskip
\dssectionheader{Collection Process}

\dsquestionex{How was the data associated with each instance acquired?}{Was the data directly observable (e.g., raw text, movie ratings), reported by subjects (e.g., survey responses), or indirectly inferred/derived from other data (e.g., part-of-speech tags, model-based guesses for age or language)? If data was reported by subjects or indirectly inferred/derived from other data, was the data validated/verified? If so, please describe how.}

\dsanswer{The data was directly observable as raw text, except for the labels and the specific analysis, which were annotated and extracted manually. The data was collected from \thebook.
}

\dsquestionex{What mechanisms or procedures were used to collect the data (e.g., hardware apparatus or sensor, manual human curation, software program, software API)?}{How were these mechanisms or procedures validated?}

\dsanswer{The data was gathered by automatically parsing the book \thebook\ through the python library \textit{fitz}\footnote{\url{https://github.com/pymupdf/PyMuPDF}}. The separation could mostly be done through rule based parsing. Only the labels, and the specific analyses were annotated manually. Correctness of the data parsing method was validated manually.
}
 
\dsquestion{If the dataset is a sample from a larger set, what was the sampling strategy (e.g., deterministic, probabilistic with specific sampling probabilities)?}

\dsanswer{N/A.
}

\dsquestion{Who was involved in the data collection process (e.g., students, crowdworkers, contractors) and how were they compensated (e.g., how much were crowdworkers paid)?}

\dsanswer{Unknown to the authors of the datasheet.
}

\dsquestionex{Over what timeframe was the data collected? Does this timeframe match the creation timeframe of the data associated with the instances (e.g., recent crawl of old news articles)?}{If not, please describe the timeframe in which the data associated with the instances was created.}

\dsanswer{Unknown to the authors of the datasheet.
}

\dsquestionex{Were any ethical review processes conducted (e.g., by an institutional review board)?}{If so, please provide a description of these review processes, including the outcomes, as well as a link or other access point to any supporting documentation.}

\dsanswer{Unknown to the authors of the datasheet.
}

\dsquestionex{Does the dataset relate to people?}{If not, you may skip the remaining questions in this section.}

\dsanswer{Unknown to the authors of the datasheet.
}

\dsquestion{Did you collect the data from the individuals in question directly, or obtain it via third parties or other sources (e.g., websites)?}

\dsanswer{N/A.
}

\dsquestionex{Were the individuals in question notified about the data collection?}{If so, please describe (or show with screenshots or other information) how notice was provided, and provide a link or other access point to, or otherwise reproduce, the exact language of the notification itself.}

\dsanswer{Unknown to the authors of the datasheet.
}

\dsquestionex{Did the individuals in question consent to the collection and use of their data?}{If so, please describe (or show with screenshots or other information) how consent was requested and provided, and provide a link or other access point to, or otherwise reproduce, the exact language to which the individuals consented.}

\dsanswer{Unknown to the authors of the datasheet.
}

\dsquestionex{If consent was obtained, were the consenting individuals provided with a mechanism to revoke their consent in the future or for certain uses?}{If so, please provide a description, as well as a link or other access point to the mechanism (if appropriate).}

\dsanswer{Unknown to the authors of the datasheet.
}

\dsquestionex{Has an analysis of the potential impact of the dataset and its use on data subjects (e.g., a data protection impact analysis) been conducted?}{If so, please provide a description of this analysis, including the outcomes, as well as a link or other access point to any supporting documentation.}

\dsanswer{No.
}

\dsquestion{Any other comments?}

\dsanswer{None.
}

\bigskip
\dssectionheader{Preprocessing/cleaning/labeling}

\dsquestionex{Was any preprocessing/cleaning/labeling of the data done (e.g., discretization or bucketing, tokenization, part-of-speech tagging, SIFT feature extraction, removal of instances, processing of missing values)?}{If so, please provide a description. If not, you may skip the remainder of the questions in this section.}
\dsanswer{Instances in which the correct answer refers to other answers (e.g. "Answer C and D are correct" were removed. For these instances, the solution label was adjusted such that the two answers were labeled as correct.
}

\dsquestionex{Was the “raw” data saved in addition to the preprocessed/cleaned/labeled data (e.g., to support unanticipated future uses)?}{If so, please provide a link or other access point to the “raw” data.}

\dsanswer{No.
}

\dsquestionex{Is the software used to preprocess/clean/label the instances available?}{If so, please provide a link or other access point.}
\dsanswer{ \anonymity{\url{https://github.com/trusthlt/legal-argument-reasoning-task}.}
}

\dsquestion{Any other comments?}

\dsanswer{None.
}

\bigskip
\dssectionheader{Uses}

\dsquestionex{Has the dataset been used for any tasks already?}{If so, please provide a description.}

\dsanswer{No.
}

\dsquestionex{Is there a repository that links to any or all papers or systems that use the dataset?}{If so, please provide a link or other access point.}

\dsanswer{No.
}

\dsquestion{What (other) tasks could the dataset be used for?}

\dsanswer{The dataset can be used for any NLP research related to civil procedure. For example, the provided answer analysis could allow natural language generation models to automatically generate an analysis.
}

\dsquestionex{Is there anything about the composition of the dataset or the way it was collected and preprocessed/cleaned/labeled that might impact future uses?}{For example, is there anything that a future user might need to know to avoid uses that could result in unfair treatment of individuals or groups (e.g., stereotyping, quality of service issues) or other undesirable harms (e.g., financial harms, legal risks) If so, please provide a description. Is there anything a future user could do to mitigate these undesirable harms?}

\dsanswer{There is no risk.
}

\dsquestionex{Are there tasks for which the dataset should not be used?}{If so, please provide a description.}

\dsanswer{We advocate using the dataset for tasks in the legal domain because the linguistic properties in Legal NLP may differ slightly from the general domain of argumentation and reasoning. 
Please also note the \hyperlink{label:termsofuse}{terms of use}.
}

\dsquestion{Any other comments?}

\dsanswer{None.
}

\bigskip
\dssectionheader{Distribution}

\dsquestionex{Will the dataset be distributed to third parties outside of the entity (e.g., company, institution, organization) on behalf of which the dataset was created?}{If so, please provide a description.}

\dsanswer{Yes, the dataset will be available for non-commercial research purposes only for three years beginning July 1, 2022.
}

\dsquestionex{How will the dataset will be distributed (e.g., tarball on website, API, GitHub)}{Does the dataset have a digital object identifier (DOI)?}

\dsanswer{The dataset can be obtained by contacting \anonymity{ivan.habernal@tu-darmstadt.de.}}. There is no DOI.

\dsquestion{When will the dataset be distributed?}

\dsanswer{The dataset was first released at [to be updated upon paper acceptance and publication].
}

\dsquestionex{Will the dataset be distributed under a copyright or other intellectual property (IP) license, and/or under applicable terms of use (ToU)?}{If so, please describe this license and/or ToU, and provide a link or other access point to, or otherwise reproduce, any relevant licensing terms or ToU, as well as any fees associated with these restrictions.}

\dsanswer{The parsed data copyright belongs to the author of the book \thebook. \hypertarget{label:termsofuse}{The corpus can only be used under the folowing conditions:}
1. The dataset gathered is used only for the purpose of Natural Language Processing (NLP) experiments with the aim to enhance legal NLP models and show their current incapability of reasoning (and not, under any circumstances, for commercial purposes).
2. The dataset may not be distributed further and must be deleted after completing the experiments.
3. For each publication based on the dataset, credit will be given to the author of the book and the publisher.
}

\dsquestionex{Have any third parties imposed IP-based or other restrictions on the data associated with the instances?}{If so, please describe these restrictions, and provide a link or other access point to, or otherwise reproduce, any relevant licensing terms, as well as any fees associated with these restrictions.}

\dsanswer{No.
}

\dsquestionex{Do any export controls or other regulatory restrictions apply to the dataset or to individual instances?}{If so, please describe these restrictions, and provide a link or other access point to, or otherwise reproduce, any supporting documentation.}

\dsanswer{No.
}

\dsquestion{Any other comments?}

\dsanswer{None.
}

\bigskip
\dssectionheader{Maintenance}

\dsquestion{Who will be supporting/hosting/maintaining the dataset?}

\dsanswer{\anonymity{Ivan Habernal} is supporting/hosting the dataset.
}

\dsquestion{How can the owner/curator/manager of the dataset be contacted (e.g., email address)?}

\dsanswer{\anonymity{ivan.habernal@tu-darmstadt.de}}

\dsquestionex{Is there an erratum?}{If so, please provide a link or other access point.}

\dsanswer{No.
}

\dsquestionex{Will the dataset be updated (e.g., to correct labeling errors, add new instances, delete instances)?}{If so, please describe how often, by whom, and how updates will be communicated to users (e.g., mailing list, GitHub)?}

\dsanswer{No substantial updates are planned, however, we will fix bugs if any are reported and communicate accordingly through the standard channels (e.g., GitHub, Twitter).}

\dsquestionex{If the dataset relates to people, are there applicable limits on the retention of the data associated with the instances (e.g., were individuals in question told that their data would be retained for a fixed period of time and then deleted)?}{If so, please describe these limits and explain how they will be enforced.}

\dsanswer{No.
}

\dsquestionex{Will older versions of the dataset continue to be supported/hosted/maintained?}{If so, please describe how. If not, please describe how its obsolescence will be communicated to users.}

\dsanswer{No.
}

\dsquestionex{If others want to extend/augment/build on/contribute to the dataset, is there a mechanism for them to do so?}{If so, please provide a description. Will these contributions be validated/verified? If so, please describe how. If not, why not? Is there a process for communicating/distributing these contributions to other users? If so, please provide a description.}

\dsanswer{No.
}

\dsquestion{Any other comments?}

\dsanswer{None.
}

\end{document}